\documentclass[letterpaper, 10 pt, conference]{ieeeconf}  

\IEEEoverridecommandlockouts                              

\usepackage{graphicx}
\usepackage{bm}
\usepackage{color}
\usepackage{hyperref}
\usepackage{multirow}
\usepackage{multicol}
\usepackage{changepage}

\usepackage{subcaption}
\usepackage{mathtools}
\usepackage{gensymb}
\usepackage{amsmath}
\usepackage{blindtext, graphicx}
\usepackage{multicol}
\usepackage{cite}
\usepackage{longtable,tabularx}
\usepackage{color,colortbl}
\usepackage{textcomp,gensymb}
\usepackage{algpseudocode}
\usepackage{algorithm}
\usepackage{caption}

\title{\LARGE \bf Neural Network-PSO-based Velocity Control Algorithm \\for Landing UAVs on a Boat}

\author{ Li-Fan Wu, Zihan Wang, Mo Rastgaar, and~Nina Mahmoudian$^{1}$
\thanks{$^{1}$Li-Fan Wu, Zihan Wang, and Dr. Mo Rastggar, and Dr. Nina Mahmoudian are with the school of Mechanical Engineering, Purdue University, West Lafayette, IN, USA.
         {\tt  wu1714, wang5044, rastgaar, ninam@purdue.edu}}%
}

\begin{document}
\maketitle
\thispagestyle{empty}
\pagestyle{empty}

\begin{abstract}

Precise landing of Unmanned Aerial Vehicles (UAVs) onto moving platforms like Autonomous Surface Vehicles (ASVs) is both important and challenging, especially in GPS-denied environments, for collaborative navigation of heterogeneous vehicles.
UAVs need to land within a confined space onboard ASV to get energy replenishment, while ASV is subject to translational and rotational disturbances due to wind and water flow. Current solutions either rely on high-level waypoint navigation, which struggles to robustly land on varied-speed targets, or necessitate laborious manual tuning of controller parameters, and expensive sensors for target localization. Therefore, we propose an adaptive velocity control algorithm that leverages Particle Swarm Optimization (PSO) and Neural Network (NN) to optimize PID parameters across varying flight altitudes and distinct speeds of a moving boat. The cost function of PSO includes the status change rates of UAV and proximity to the target. The NN further interpolates the PSO-founded PID parameters. The proposed method implemented on a water strider hexacopter design, not only ensures accuracy but also increases robustness. 
Moreover, this NN-PSO can be readily adapted to suit various mission requirements. Its ability to achieve precise landings extends its applicability to scenarios, including but not limited to rescue missions, package deliveries, and workspace inspections. \href{https://youtu.be/6NSpt4S8kcs}{[Video]}

\end{abstract}

\begin{keywords}
Aerial Systems: Perception and Autonomy;
Machine Learning for Robot Control.

\end{keywords}


\section{Introduction} %

Many researchers equip drones with expensive sensors like depth cameras, dual cameras\cite{falanga2017vision, du2020real}, LiDAR\cite{kim2017lidar}, or even high-cost IR-Lock sensors to deal with the difficulties of precise landing. Although these added sensors bring benefits to a more precise landing, their weight and power consumption can not be ignored due to the limited battery life of drones. In this work, the landing process only depends on a low-cost RGB camera to detect fiducial markers and estimate the landing target location, plus a distance sensor to detect the height of the target plane.

In addition, software algorithms have been developed for the precise landing of drones. Advances in drone state estimator and visual inertia odometry\cite{santamaria2022towards, qin2018vins, campos2021orb, 5174709} have significantly improved trajectory following accuracy and stability in waypoint navigation tasks. However, for safe and soft drone landing, regular aerial waypoint navigation may not be directly adapted due to the ground effect when the drone touches the landing platform. Besides, waypoint navigation dependency on drone odometry brings another source of error in localizing dynamic landing targets, compared to direct vision-driven velocity controller which does not focus on following certain trajectories when landing on a dynamic target. For example, when landing on a moving boat, forest canopies decrease the GPS accuracy. The sunlight reflection, shadow, and uneven and low-textured lake surface disturb the visual odometry positioning\cite{aqel2016review}. To ensure the robustness of landing in GPS-denied river environments, our work adopts an intermediate-level velocity controller.

To achieve a velocity controller for fast, stable, and accurate horizontal alignment with the landing target, many researchers investigate the fuzzy logic \cite{bouaiss2023visual}, interaction matrix \cite{keipour2022visual, li2021image, drones7020130}, model prediction \cite{feng2018autonomous, BEREZA202015180} or the cascade method \cite{ZHAO20212301}. 
However, many of these approaches require either manual parameter tuning before deployment (For example, fine-tuning multi PID parameters for different heights \cite{Lee2020AVC}) or have requirements about the differentiability of the optimization function (such as adjusting PID values with exponential functions based on heuristic rules \cite{app9132661,s23020829, ZHAO20221}).

To take into account both the accuracy and smoothness of drone landing with minimum human intervention and localization requirement, an adaptive velocity controller is proposed and optimized by NN-PSO algorithm. The controller includes variable PID parameters for different altitudes and boat speeds. Besides, an exponentially decaying function (\textit{tanh} in this work) is adopted to regulate z-direction velocity control for stable landing with braking effects \cite{gautam2022autonomous, shi2019neural}. By taking advantage of PSO's exploration \cite{mac2016ar} and NN's expansion \cite{robotics7040071,choi2021robust}, the method can keep the accuracy and greatly increase the training efficiency.
Lastly, field experiments were conducted to test the sim-to-real transferability of the velocity controller that lands a hexacopter on a boat. The qualitative result verifies the precision and stability, even in windy conditions, of the NN-PSO-based velocity control algorithm. Low-cost dependent sensors and negligible parameter tuning effort from simulation training to real-world deployment reveal the wide range of applicability and simplicity of the proposed drone landing approach.


This paper is organized in the following way:  Sec.~\ref{sec:Algorithm} elaborates on the NN-PSO inputs, the cost function design, the adaptive velocity controller, target state estimation, and state machine. Sec.~\ref{sec:exper} presents the water strider drone design, the training process of PSO particles, the comparison of the constant PID and the adaptive PID controllers, the velocity ramp function effectiveness, the comparison of single and dual marker design, and the landing accuracy in different environments. The conclusions are given in Sec.~\ref{sec:conc_future}.

\section{Algorithm and Controller} \label{sec:Algorithm}

This section first elaborates on the inputs, cost function design, and particle moving rule of the PSO algorithm. Second, the adaptive PID controller, the ramp function and \textit{tanh} function are adopted for three-dimensional speed control. Third, the inputs and outputs of the Neural Network are explained. Fourth, the state estimation error of the landing target is reduced by filtering. Fifth, the landing strategy, including three stages: Explore, Align, and Land, is depicted.

\subsection{Particle Swarm Optimization}

PSO algorithm is a population-based optimization algorithm that is inspired by the behavior of social animals, such as birds and fish. With each particle being a set of parameters that needs to be optimized, PSO iteratively updates the velocity and position of the particles in the parameter search space to find the optimal solution. In this work, PSO helps optimize the parameters of the velocity controllers to ensure the accurate and stable landing of drones on the targets. 

The proposed velocity controller includes proportional, integral, and derivative parameters for the horizontal alignment, and the scaled \textit{tanh} function for stable descending with brake. Thus, each particle includes five parameters.

\begin{equation} \label{equ:X_input}
    X^i(t) = (K_P^i(t),K_I^i(t), K_D^i(t),\alpha^i(t),\beta^i(t))
\end{equation}
where the position of particle $i$ at iteration $t$ is denoted as $X^i(t)$. $K_P$, $K_I$, and $K_D$ represent the PID parameters of the drone's horizontal velocity controller. $\alpha$ is the scale rate of the vertical velocity controller, and $\beta$ is the ramp function time, which will be elaborated in Sec.~\ref{subsec:vel_control}.

First, the basic optimization function is the horizontal distance between the target location and the final landing spot of the drone. 
\begin{equation} \label{equ:d(x)}
    \begin{split}
    d =  \sqrt{(x_{d}-x_{t})^2+(y_{d}-y_{t})^2} 
    \end{split}
\end{equation}
$x_{d}$ and $y_{d}$ are the final position where the drone actually land, $x_{t}$ and $y_{t}$ are the position where the target is. Therefore, if $d$ is very close to 0, it represents that the input, PID parameters, are suitable and help the drone land precisely.

Second, to optimize the vertical landing speed and ensure stability, the optimization function further includes time, velocity, acceleration, and the roll and pitch rate. The goal of the PSO is to find the minimum of the function.   
\begin{equation} \label{equ:cost}
    \begin{split}
    F(X) = d+T+\left|\nu\right|+\left|a_z\right|+\sum\left|{\dot{\phi}}\right| +\sum\left|{\dot{\theta}}\right|
    \end{split}
\end{equation}
where $T$ is the duration from when the drone hovers vertically over the target, to the drone landing on the surface. $\nu$ and $a_z$ are the z-directional velocity and acceleration at the moment when the drone reaches the surface. $\dot{\phi}$ and $\dot{\theta}$ are the x-axis and y-axis angular velocity, which represent the roll and pitch rate of the UAV. 


Third, each particle $i$ will update its own velocity vector $V$ in every iteration $t$. The velocity vector of particle $i$ updates towards not only this particle's best parameter position but also the best global parameter position among all spread particles.
\begin{equation} \label{equ:pso_vel}
    \begin{split}
    X^i(t+1) = X^i(t) + V^i(t+1) 
    \end{split}
\end{equation}
\begin{equation}\label{equ:pso}
    \begin{split}
        V^i(t+1) = wV^i(t) + c_1 r_1 (pbest^i - X^i(t)) \\ 
                           + c_2 r_2 (gbest - X^i(t))
    \end{split}
\end{equation}
where velocity vector at next iteration $V^i(t+1)$ is the weighted sum of the current velocity vector $V^i(t)$, vector to current particle's best parameter $pbest^i$ and vector to the best parameter among all particles $gbest$, with weights being $w$, $c_1$, and $c_2$ respectively. Weight $w$ acts as an inertia force to particles to avoid sudden parameter jump that jeopardizes convergence, weights $c_1$, and $c_2$ improve the chance of finding the global sub-optimal solution without being stuck in the local optimal region. $r_1$ and $r_2$ are random scalars ranging [0, 1). They add noise to the velocity vector magnitude to further increase the convergence rate \cite{engelbrecht2007computational}.

Last, the PSO algorithm can efficiently explore the sub-optimal global solution. Every particle represents a set of PID parameters that govern the horizontal movements of a drone during a complete landing process.
In each iteration, particles share information on costs according to Eqn.~\ref{equ:cost} and update their velocity vectors based on this collective knowledge. This approach makes the discovery of near-optimal velocity controller parameters possible with less than a hundred iterations in simulation. 
Besides, the hyper-parameters in Eqn.~\ref{equ:pso} can adopt task-independent empirical values with little manual tuning. 
Moreover, the optimization of the cost function can be easily customized without the need to consider its differentiability with respect to the parameter variables. This enables researchers to mainly focus on the parameter design of velocity controllers and task-dependent cost functions.

\subsection{Adaptive Velocity Controller} \label{subsec:vel_control}

The PID controller is adopted for horizontal control, as the below equation:
\begin{equation} \label{equ:PID_discrete}
    \begin{split}
    v_{xy}= K_P*e + K_I*(I_e+e*T)
                  + K_D*\frac{e-e'}{T}
    \end{split}
\end{equation}
where $K_P$ is the proportional term, $K_I$ is the integral term, $I_e$ is the accumulated error, $K_D$ is the derivative term, $e$ is the current error, $e'$ is the last error, and $T$ is the sampling period, which is 0.05 sec in this experiment. 
  
\begin{figure}[htp]
    \centering
    \includegraphics[width=1\linewidth]{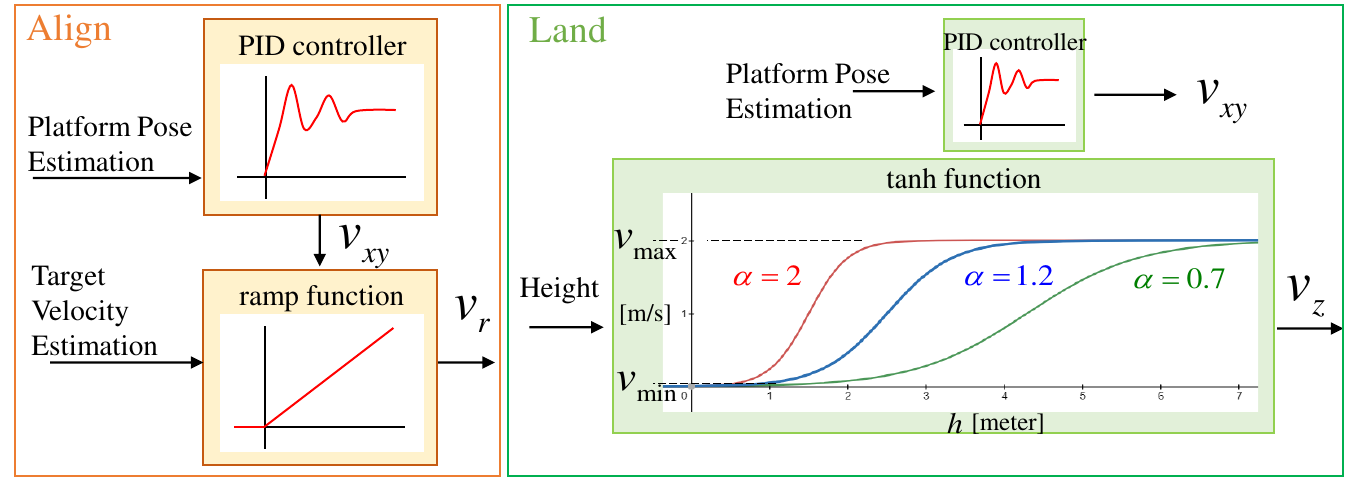}
    \caption{From Align to Land stage, velocity controllers change}
    \label{fig:controller_change}
\end{figure}

To optimize the landing stability and time spent, the vertical velocity controller adopts the \textit{tanh} function, which can make the drone descend fast at high altitudes and slow down at low altitudes. 
\begin{equation} \label{equ:tanh}
   v_{z} = \frac{v_{max}}{2}(\tanh{(\alpha h-3)} +1)+v_{min}
\end{equation}
where the scaling parameter $\alpha$ can be optimized to change the slope of $v_z$ to exponentially regulate drone descending velocity at various altitudes in Fig.~\ref{fig:controller_change}. $h$ is the real-time height in meters measured by the range sensor. $v_{max}$ and $v_{min}$ are the maximum and minimum vertical velocity. The constants in the equation is for shifting the tanh function towards within the bounds of positive $v_{max}$ and $v_{min}$ values. 
To track a moving boat effectively, a ramp velocity function is implemented to work in tandem with the PID controller during the Align stage. This harmonious integration enables the drone to gradually increase its velocity in a smooth manner. Maintaining a stable acceleration profile allows to reduce oscillations and maintain steady roll and pitch angles, which are instrumental in enhancing the accuracy of boat tracking by keeping it within the camera field of view.
\begin{equation}
\begin{aligned}
v_{r} &= f(v_{xy}, \beta) = \\
&\begin{cases}
    v_{r}+\lambda & \text{if } v_{xy} > v_{r}+\delta \text{ and }t<\beta \\
    v_{r}-\lambda & \text{if } v_{xy} < v_{r}+\delta \text{ and }t<\beta \\
    v_{xy} & \text{else}
\end{cases}
\end{aligned}
\end{equation}
where $v_{r}$ is the output velocity of the ramp function. $\lambda$ is the desired acceleration. The velocity threshold $\delta$ and the time threshold $\beta$ decide whether to directly use the PID controller output $v_{xy}$. When tracking a fast-moving target, the integrator requires more time (higher $\beta$) to eliminate steady-state error.

\subsection{Neural Network} \label{subsec:NN}

As a drone transitions from higher to lower altitudes, its field of vision narrows, prompting the need for increased $K_p$ values to enhance horizontal agility and target tracking. Moreover, the varying speed of a moving boat also requires different PID parameters for velocity compensation. The studies referenced in \cite{Lee2020AVC,app9132661,s23020829} underscore the critical relationship between flight altitudes and specific PID parameter adjustments. The application of a Neural Network serves to provide a more comprehensive solution for adapting PID parameters across various scenarios, effectively reducing the reliance on an extensive number of PSO training cases.

\begin{figure}
    \centering
    \includegraphics[width=\linewidth]{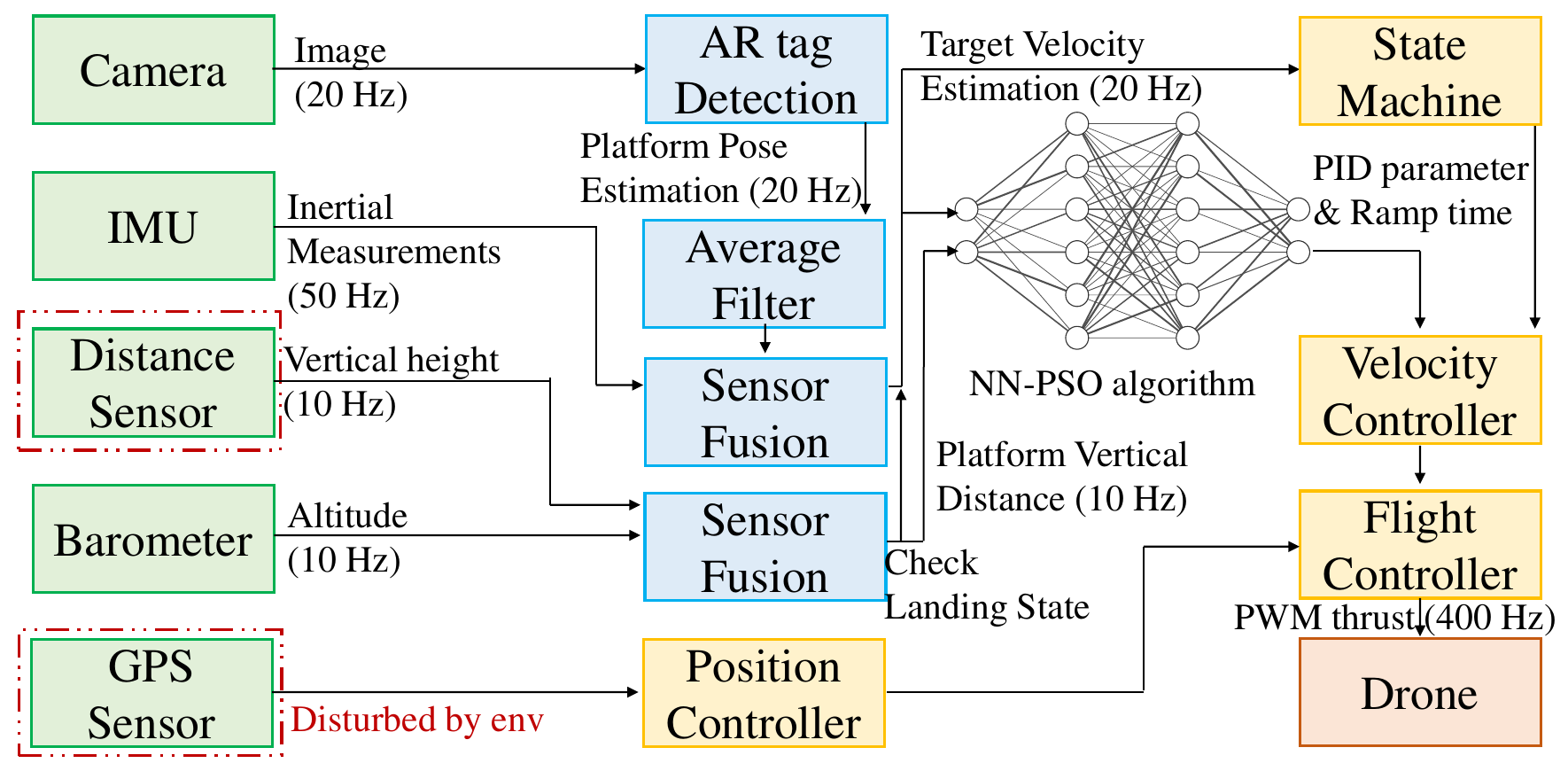}
    \caption{The system overview: Yellow blocks are the controllers; Green blocks are sensor modules; and the blue blocks represent software design. The communication between modules is through ROS.}
    \label{fig:framework}
    \vspace{-4 mm}
\end{figure}\textbf{}
The initial phase involves training the neural network using a flexible dataset of 25 data points, which can be tailored to user requirements. This dataset includes flight height and boat velocity as inputs, while PID parameters and the ramp function time $\beta$ serve as the corresponding outputs.

These 25 training data points represent the outcomes produced by PSO, covering a range of five distinct flight altitudes and five different boat speeds. To provide context, the maximum observable height of the drone is limited to 5 m and the altitude of the landing platform on the boat is about 1.25 m. The maximum speed of the boat is 2.9 m/s. Accordingly, the five flight altitudes are set between 1.3 and 5 m, while boat speeds span from 0 to 2.9 m/s.

Leveraging the interpolation capabilities of the Neural Network, we can construct a dynamic PID controller that adapts seamlessly to varying heights and boat speeds. During field tests, the drone can readily utilize this Neural Network by inputting observed boat velocities, allowing it to adjust PID parameters in real-time according to its current altitude. This process is visualized in Fig.~\ref{fig:framework}.

\subsection{Target State Estimation} \label{subsec:target_state}
Drones typically utilize pitch or roll angles to adjust flight direction, which in turn affects the camera orientation on the drone. To correct for this observational error, transformation matrices are employed to rotate coordinates.

\begin{equation} \label{equ:state_estimation_2}
   P_{\text{body}} =  T_{\text{body\_world}} T_{\text{world\_cam}} P_{\text{cam}} = T_{\text{body\_cam}} P_{\text{cam}} 
\end{equation}
where real-time roll ($\phi$) and pitch ($\theta$) angles of the drone can be directly obtained from the IMU sensor. When projecting the target position onto the camera frame, only $R_x$ and $R_y$ are needed, so the yaw ($\psi$) angle is set as 0. 
Last, the movement of the target in the drone body frame, divided by the time can derive the velocity estimation:
\begin{equation} \label{equ:velocity_estimation}
   \overline{\dot{x(t)}} =  (\sum_{k=t-n}^{t} \frac{\Delta{x(k)}}{T_s})/n
\end{equation}
where a filter samples and averages $n$ velocities to reduce errors, with $x$ representing the position of the boat in the drone body frame and $T_s$ denoting the sampling time, which dynamically adjusts while adhering to an outlier threshold, because, during field tests, variations in the on-board computer computational load impact the camera sampling rate.

\subsection{Landing Strategy} \label{subsec:strategy}
\begin{figure}[htp]
    \centering
    \includegraphics[width=0.9\linewidth]{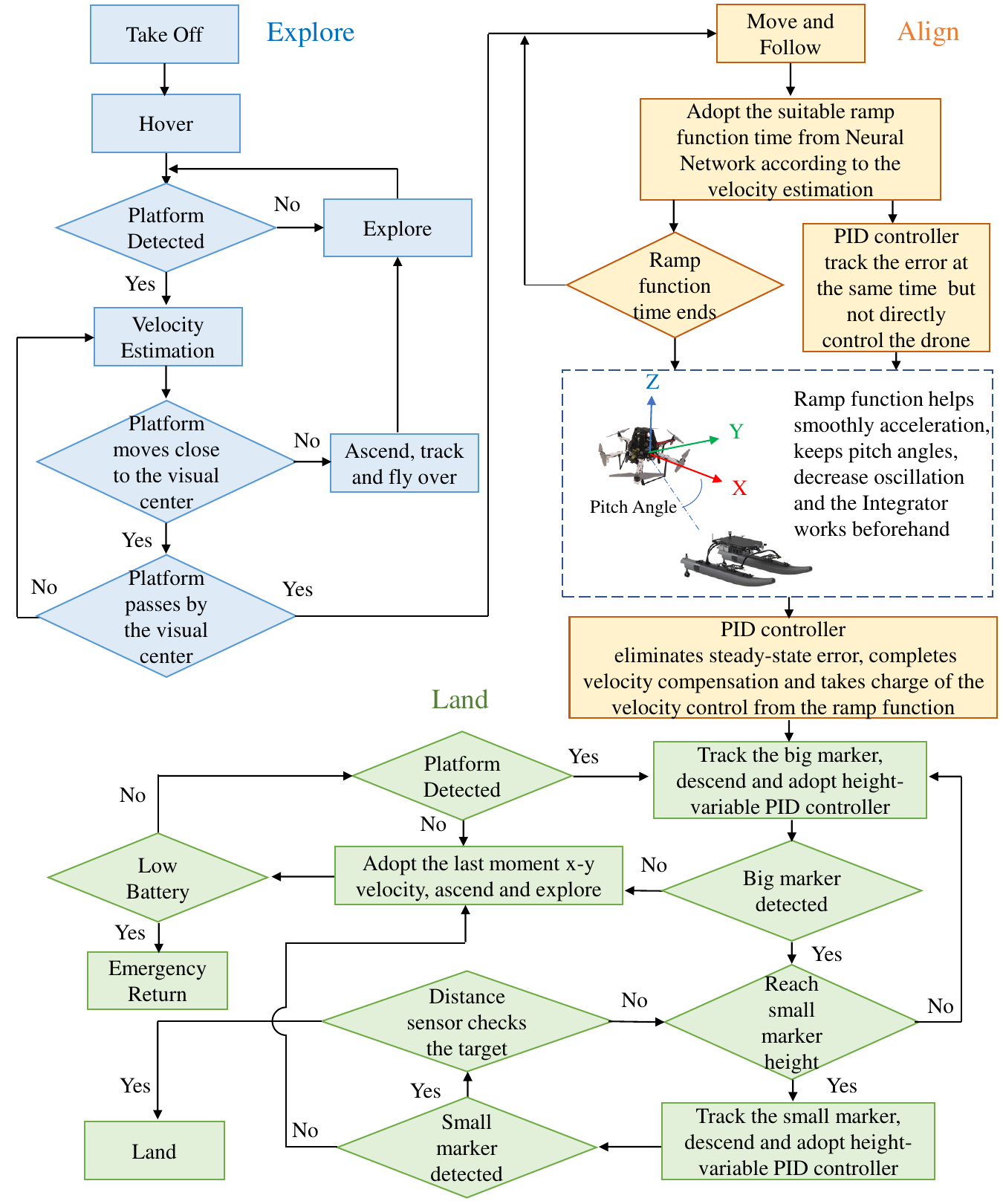}
    \caption{The flowchart of the landing strategy}
    \label{fig:flow_chart}
    \vspace{-6 mm}
\end{figure}

To ensure comprehensive coverage of various scenarios, such as missing targets and targets with varying speeds, multiple events have been integrated into the system to enhance its robustness. The mission consists of three main stages: exploration, alignment, and landing. Fig.~\ref{fig:flow_chart} illustrates these stages along with their corresponding trigger conditions.

\textbf{}

\section{Results and Analysis} \label{sec:exper} 
\vspace{-1 mm}
This section details the training and validation processes of the velocity controller for drone landing in simulation, the real-world tests of the trained controller, as well as the results and their analysis.

\subsection{Water Strider Design} \label{sec:drone_design}
\begin{figure}[h]
    \centering
    \includegraphics[width=0.9\linewidth]{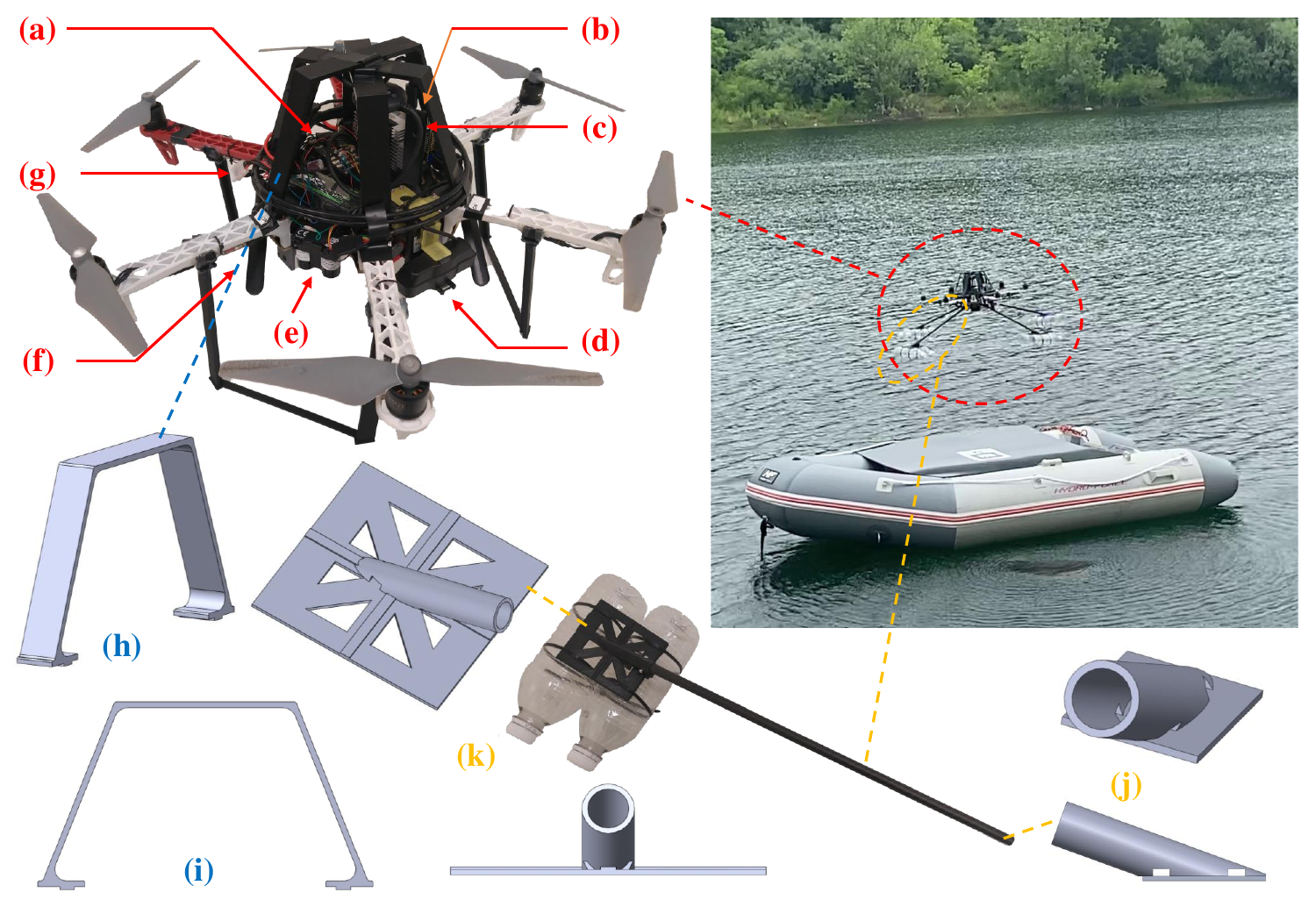}
    \caption{The hexacopter schematics, water strider design, and the boat in the experiments.}
    \label{fig:drone}
    \vspace{-6 mm}
\end{figure}

\vspace{-1 mm}
The hexacopter is inspired by water striders, featuring four legs to handle challenging scenarios like landing on a moving boat and emergency lake landings. Fig.~\ref{fig:drone} provides an overview of the autonomous drone equipped with (a) Pixhawk4 FMUv5, (b) Jetson TX1 running Ubuntu 18.04 with ROS Melodic, (c) GPS, (d) AWC201-B RGB webcam, and (e) LiDAR-Lite v3 distance sensor. Separate batteries (11.1V Li-Po) power Pixhawk4 and six 960 KV motors.
To ensure smooth operations, the distance sensor is wired with a capacitor to prevent voltage fluctuations and communicates with Pixhawk through the I2C protocol. Additionally, (f) 2.4GHz and 5GHz WiFi antennas are integrated into Jetson TX1 for wireless SSH access and off-board control. Other components include (g) Holybro Radio Telemetry and FrSky X8R Telemetry systems facilitate communication with QGroundcontrol and an RC transmitter, allowing manual flight mode. Three ABS elastic ribs, shown in (h) side view and (i) front view, are strategically positioned to safeguard the computer and controllers. Each water strider-inspired leg, composed of (j) the upper part connecting to the base and (k) the lower part outfitted with two 330 ml bottles for added buoyancy, ensures stability during landings. The total weight is approximately 2.4 kg.

\subsection{Training and Validation of the NN-PSO} \label{sec:training}

To train the velocity controller using the PSO algorithm, a five-particle configuration is utilized to strike a balance between training efficiency and optimal solution attainment. In the initialization phase of the velocity controller parameters, namely $K_P$, $K_I$, $K_D$, $\alpha$, and $\beta$ in Eqn.~\ref{equ:X_input}, we employed randomization within predefined ranges: [0, 2) for $K_P$, [0, 1) for $K_I$, [0, 1) for $K_D$, and [0, 3) for both $\alpha$ and $\beta$.
For the validation of the controller performance of UAV landing on a moving ASV in the same simulated environment, Gazebo is adopted and includes the river environment with a differential-driven WAM-V boat from \cite{bingham19toward}, a simulated hex-rotor firefly \cite{Furrer2016}, and a Iris drone to take off and land on the boat. To achieve robust controller parameters, the training sessions are separately conducted for descending and horizontal tracking. The descending motion, potentially conflicting with tracking during PSO training, allows the drone to land quickly for higher rewards instead of stable, extended tracking.

\begin{figure}
    \captionsetup{skip=-4pt} 
    \centering
    \includegraphics[width=\linewidth]{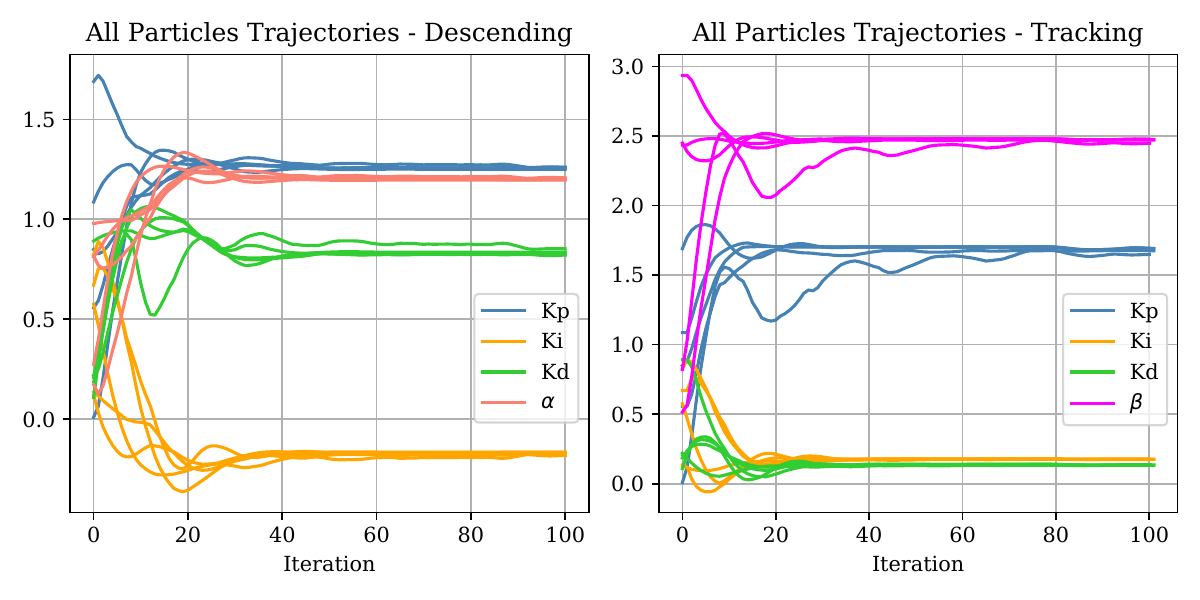}
    \caption{PID gains, $\alpha$ and $\beta$ value of the five particles. The cost function trajectory of the best particle across all iterations during PSO training.}
    \label{fig:PSO_train}
    \vspace{-2 mm}
\end{figure}

In the left Fig.~\ref{fig:PSO_train}, the trajectories of the five particles converge towards optimal descending parameters. These trajectories document the adjustments made to $K_P$, $K_I$, $K_D$, and $\alpha$. Notably, the cost of the best particle drops rapidly within the initial 20 iterations, indicating the swift convergence of the PSO algorithm towards optimal values. Subsequently, by sharing costs, all particles gradually converge towards a global sub-optimal solution, resulting in closely aligned optimized parameters among all particles.

After 100 iterations of PSO training, utilizing the best PID parameters ($K_P = 1.565$, $K_I = 0.121$, and $K_D = 0.245$) at that moment, the horizontal distance error between the landed drone and the target spot is reduced to a mere 0.0026 meters within the training simulation environment. In comparison to the default Pixhawk AutoLand Mode, descending from the same height results in a significant reduction in landing time, from 18.62 to 3.21 seconds, while the vertical velocity $\nu$ decreases from -0.3041 to 0.002 m/s. Additionally, the vertical acceleration $a_z$ experiences a negligible increase, from 9.817 to 9.821 $\mathrm{m/s^2}$
 (including gravitational acceleration). This indicates that the drone can land quickly without bouncing upon contact with the surface of a boat.

\begin{figure}
    \centering
    \captionsetup{skip=-1pt} 
    \captionof{table}{The yellow blocks are the outputs of the PSO, and the other blocks are the result of the Neural Network.}
    \includegraphics[width=\linewidth]{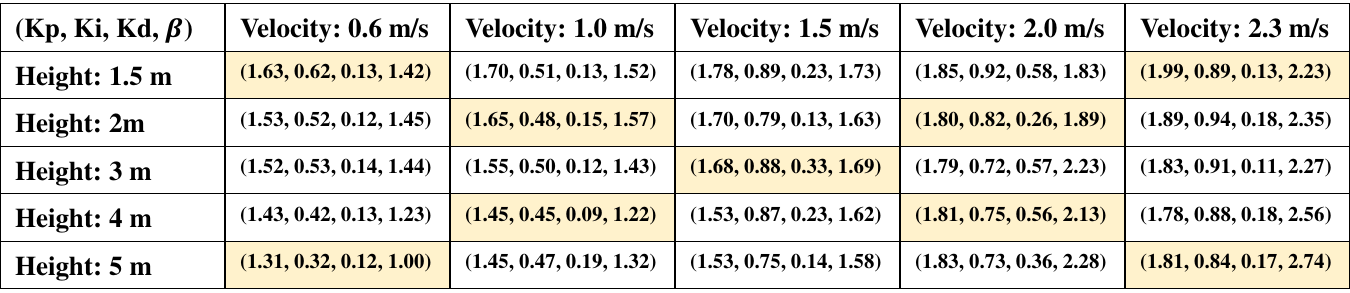}
    \label{table:NN_table}
    \vspace{-8 mm}
\end{figure}

The right Fig.~\ref{fig:PSO_train} shows the trajectories of the five particles moving close to the optimized horizontal tracking parameters. These trajectories document the adjustments made to $K_P$, $K_I$, $K_D$, and $\beta$. The PSO algorithm separately optimized these parameters at five different heights and positions, which are marked in yellow in Tab.~\ref{table:NN_table}. Subsequently, the trained neural network model predicts continuous parameters for altitudes from 1.5 to 5 m and velocity from 0.6 to 2.3 m/s.

\subsection{Adaptive PID controller}

The Adaptive PID controller exhibits an impressive maximum tracking speed of 8.9 m/s (target speed) when deployed on the Iris drone, which itself can reach a top speed of 11 m/s. Thus, the maximum landing speed achieved with this controller reaches 80.9\% of the drone maximum flight speed (the target size in the experiment is 1x1 m).

\begin{figure*}
  \centering
  \includegraphics[width=\linewidth]{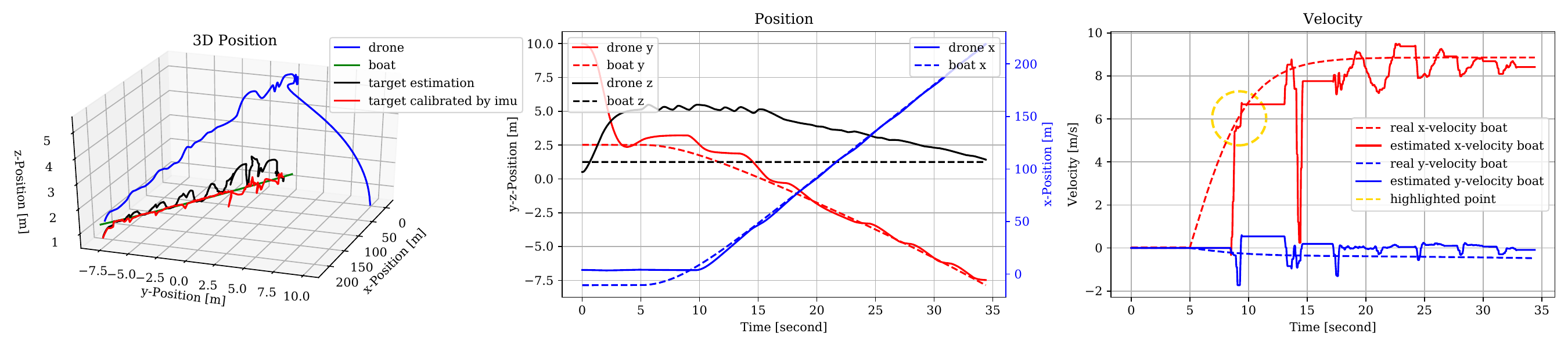}
  \caption{The simulation result: Iris drone can land on a boat that moves at a speed of 8.9 m/s.}
  \label{fig:fast_land}
  \vspace{-2 mm}
\end{figure*}

In Fig.~\ref{fig:fast_land}, the 3D and 2D trajectories of the flight, along with target velocity estimation, are depicted. At the 9-second mark in the velocity plot, the drone velocity measurement is sufficiently accurate, serving as input for the neural network in the proposed landing strategy.

\begin{figure*}
  \centering
  \includegraphics[width=\linewidth]{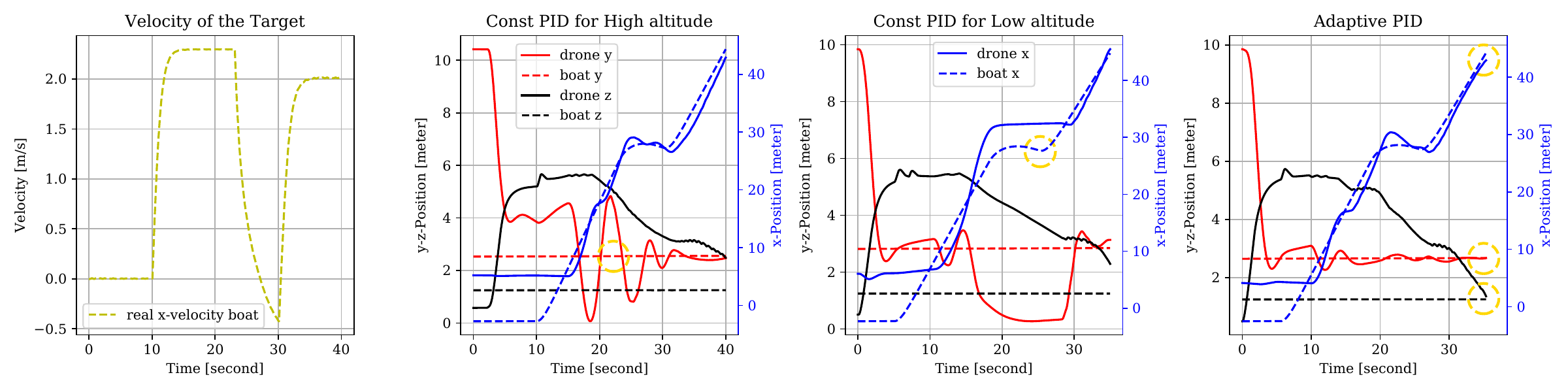}
  \caption{The results of constant and Adaptive PID comparison experiments.}
  \label{fig:const_PID}
  \vspace{-4 mm}
\end{figure*}

\begin{figure*}
  \centering
  \includegraphics[width=\linewidth]{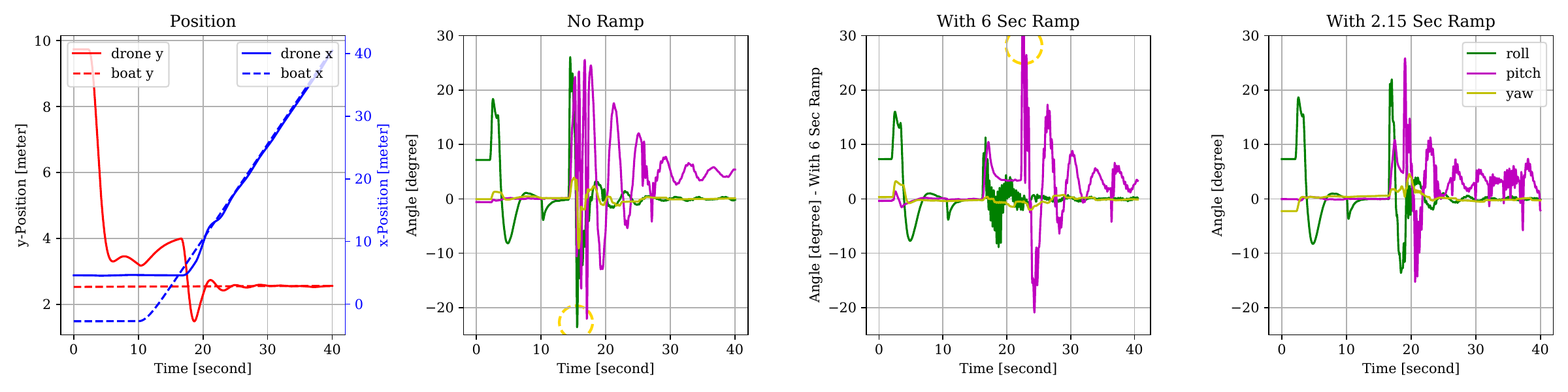}
  \caption{The results of ramp function duration comparison experiments. The appropriate 2.15 sec ramp function in 4th column, can help the drone effectively decrease oscillation and successfully track the target.}
  \label{fig:ramp}
  \vspace{-4 mm}
\end{figure*}

In comparison, the Visual Odometry Tracking \cite{falanga2017vision} exhibits a maximum tracking speed of 4.2 m/s in simulation scenarios. When deployed on the DJI F450 drone, which itself can reach a top speed of 36.1 m/s, its controllers' tracking capability is far from the hardware limit. The target landing platform size in their experiment is 1.5x1.5 m. The maximum landing speed is about 11.6\% of the drone's maximum flying speed. Hence, the utilization of the NN-PSO algorithm in conjunction with the adaptive PID controller streamlines testing and exploration, allowing us to approach the landing speed near the maximum flight speed more efficiently.

Many studies \cite{app9132661, Lee2020AVC} have demonstrated that the height-adaptive PID controller outperforms its constant counterpart. To further substantiate this theory and validate the superior performance of our Adaptive PID controller, a comparative analysis involving two constant PID controllers is executed.
These two PID controllers were fine-tuned using PSO at both low and high altitudes, serving as benchmarks. The descent speed remained constant, allowing ample time for observing the controllers' performance. Furthermore, the boat's variable speed, as opposed to a constant speed, enabled a comprehensive evaluation of the PID controllers' tracking capabilities. In Fig.~\ref{fig:const_PID}, the fourth column illustrates that the adaptive PID controller excels in both tracking ability and boat velocity estimation, attributed to its stable flight performance.

\subsection{Ramp Function Effectiveness Assessment}

The ramp function can gradually increase the drone's speed, maintains camera stability, ensures the target remains in view, and allows the integrator to proactively contribute. Once the ramp function phase concludes, the PID controller takes over, with its integrator finalizing velocity compensation. The accompanying Fig.~\ref{fig:ramp} illustrates the impact of varying ramp function duration. If the duration is too brief or omitted altogether, the drone may exhibit an initial overreaction and miss the target. Especially when the marker initially appears at the edge of the drone camera's field of view, large estimated distance typically generates substantial rush. Conversely, if the duration is excessively prolonged, the drone may struggle to land accurately during descent. By employing an appropriately timed ramp function, the drone can effectively track and execute a successful landing.

\subsection{Dual Marker Performance Evaluation}

\begin{figure}
    \centering
    \includegraphics[width=0.8\linewidth]{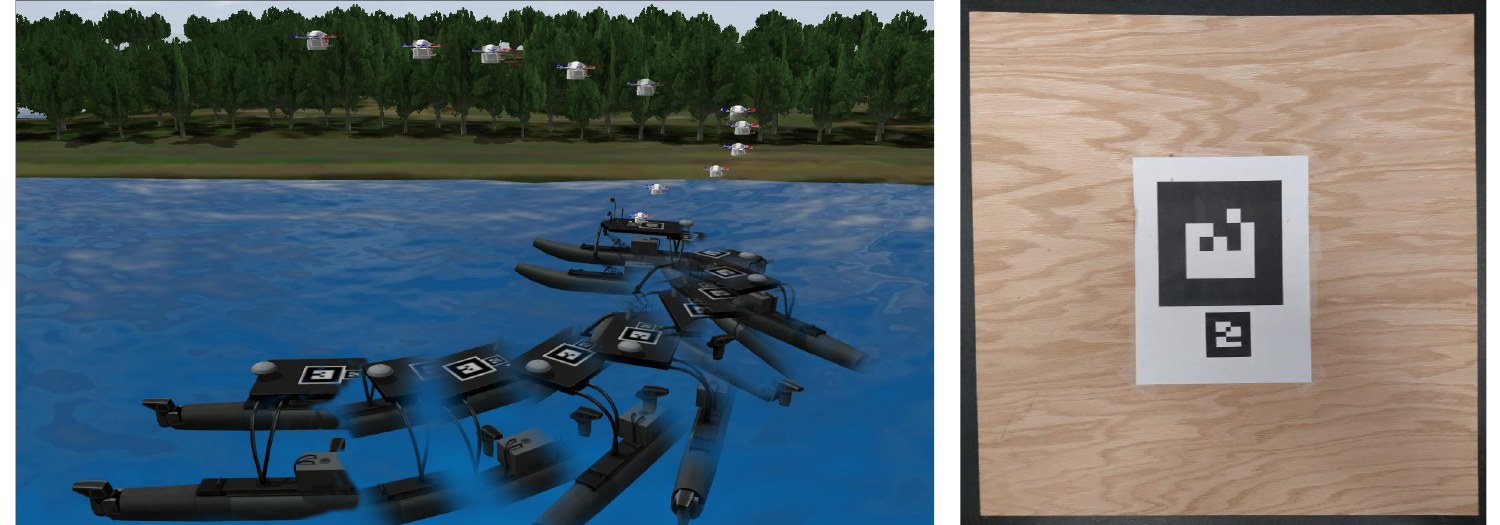}
    \caption{The circularly moving boat and the real dual marker.}
    \label{fig:marker}
    \vspace{-4 mm}
\end{figure}

In the context of tracking landing targets, ARTag \cite{fiala2005artag, 9292607, 8958701} markers, known for their binary square fiducial properties, offer a swift and robust solution for 3D camera pose estimation.
In our experimental setups, the Dual Marker is designed and attached on the boat to facilitate low-altitude target tracking and delicate flight maneuvers. The Dual Marker comprises two ARTags, one measuring 5.5x5.5 cm and the other 16x16 cm, positioned 1 cm apart.
To further validate the effectiveness of the Dual Marker, we conducted tests using a circularly moving boat scenario in Fig.~\ref{fig:marker}, and the data is recorded in Tab.~\ref{table:dual_mark}. Under identical conditions, including the same PID settings and velocity limitations, the Dual Marker design significantly improves the drone's landing precision during simultaneous forward movement and descent, thereby reducing landing position errors compared to using a single marker.
In summary, transitioning to the smaller marker when descending proves instrumental in helping the drone overcome the limited field of vision at low altitudes.

\begin{table}
\begin{center}
\caption{The distance error of the drone landing on the boat with different moving patterns}
    \begin{tabular}{c|c|c}
        \hline
        Error (m)           & Single Marker     & Dual Marker \\
        \hline
        \textbf{Static}                                           & 0.113     & 0.012 \\
        \hline
        \textbf{Linear movement}                                  &           &        \\    
        speed 0.1 m/s                                    & 0.0929    & 0.0558 \\
        speed 0.2 m/s                                    & 0.0881    & 0.0117 \\
        speed 0.4 m/s                                    & fail      & 0.2497 \\
        \hline
        \textbf{Circular movement}                                &           &        \\ 
        \textbf{with x-speed 0.1 m/s}                             &           &        \\ 
        speed 0.1 rad/s                                  & 0.1539    & 0.0781 \\
        speed 0.2 rad/s                                  & 0.1708    & 0.1162 \\
        speed 0.3 rad/s                                  & 0.3268    & 0.2082 \\
        speed 0.4 rad/s                                  & fail      & 0.2130 \\
        \hline
    \end{tabular}
    \label{table:dual_mark}
\end{center}
    \vspace{-4 mm}
\end{table}

\subsection{Field Test}
\begin{figure}
  \centering
  \begin{subfigure}{0.23\textwidth}
    \centering
    \includegraphics[width=\linewidth]{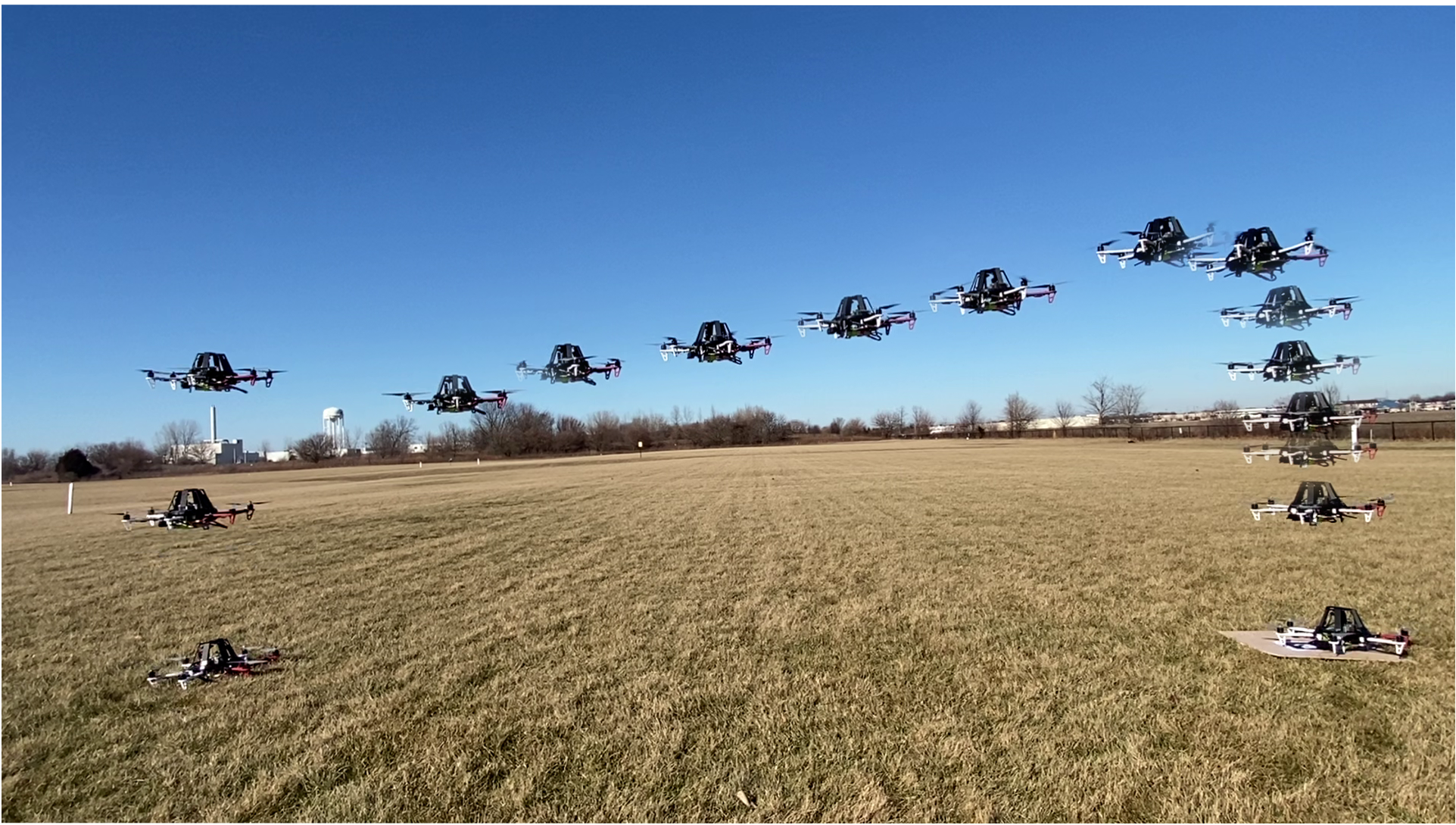}
    \label{fig:image1}
  \end{subfigure}
  \begin{subfigure}{0.23\textwidth}
    \centering
    \includegraphics[width=\linewidth]{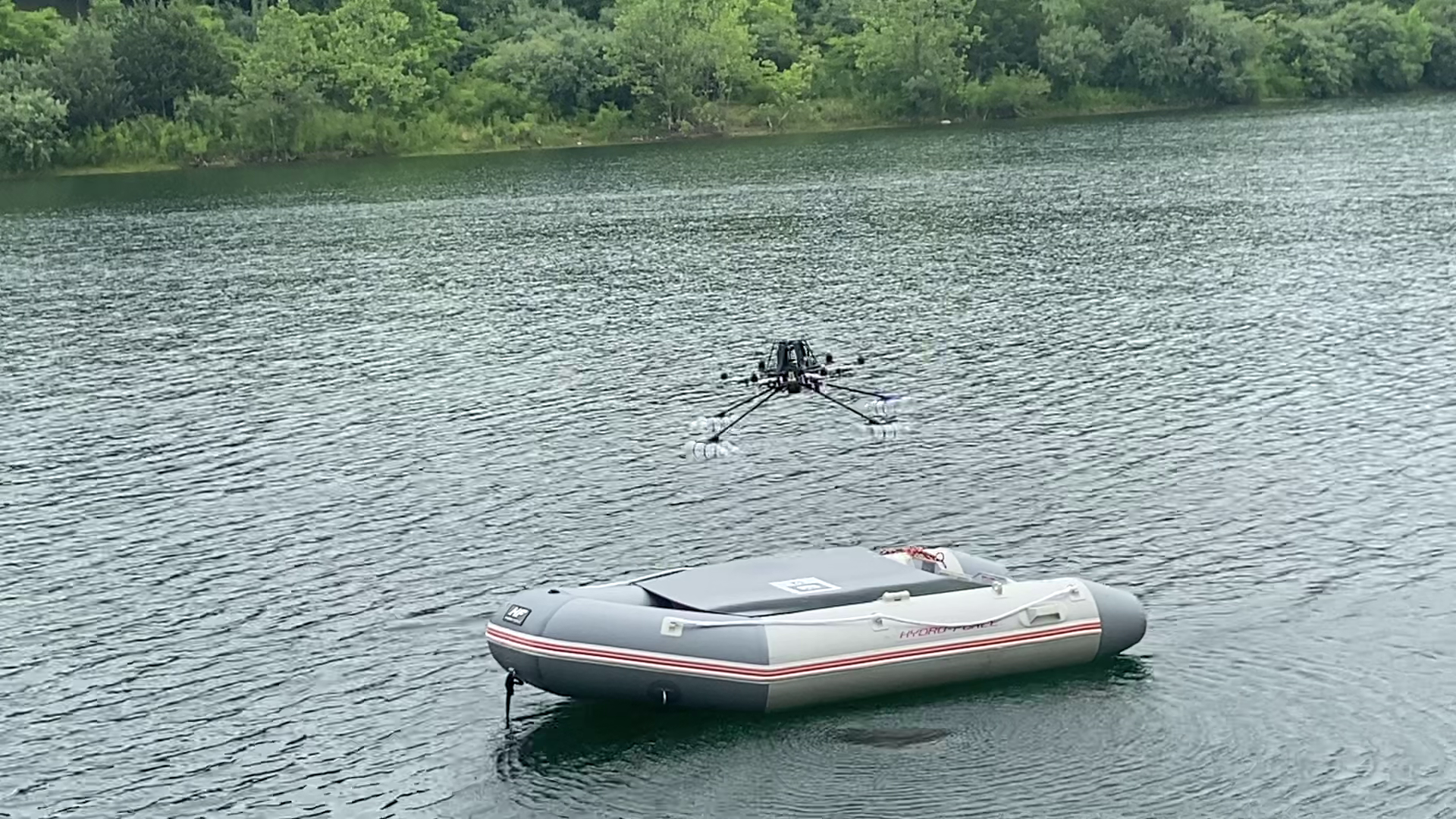}
    \label{fig:image2}
  \end{subfigure}
  \captionsetup{skip=-3pt} 
  \caption{Visual-based Drone Landing with Neural Network-PSO-based Velocity Control Algorithm}
  \label{fig:action}
  \vspace{-6 mm}
\end{figure}

The field tests demonstrate precise landing capabilities in both static and dynamic scenarios, as visualized in Fig.~\ref{fig:action}.

\section{Conclusions} \label{sec:conc_future} 
The Neural Network-PSO-based velocity control algorithm enables autonomous landing of the drone on a moving boat, using only onboard sensing and computing, eliminating the need for external infrastructure like visual odometry or GPS. The approach doesn't require prior information about the boat's location. To further increase robustness, the adaptive PID controller is designed and NN-PSO trained for variable speeds and altitudes. This algorithm includes features like a ramp function for keeping targets in vision, a tanh function for landing softly, multi-sensor fusion for relative UAV localization and boat velocity estimation. By achieving more stable flight velocity, the drone can autonomously track and land on a boat. This framework is validated through both simulation and real-world experiments.

\bibliographystyle{IEEEtran}
\bibliography{references.bib}
\end{document}